# Dual-Model Sentiment Analysis of Consumer Reviews in the Retail Coffee Sector Using Machine Learning and Deep Learning Approaches

[a]Muntasir Hasan Kanchan, [b]Md. Alamgir Hossain, [c]Md. Samiul Islam, [d]Muhammad Masud Tarek

[a, b, d]Department of Computer Science and Engineering, State University of Bangladesh, Dhaka, Bangladesh

[c]Department of Computer Science, American International University - Bangladesh, Dhaka, Bangladesh

[a, b, c]Skill Morph Research Lab., Skill Morph, Dhaka, Bangladesh

[a]E-mail: muntasirhasank@gmail.com

[b]E-mail: alamgir.cse14.just@gmail.com, ORCID: https://orcid.org/0000-0001-5120-2911

[c]E-mail: samiulislam.cse@gmail.com

[d]E-mail: tarek@sub.edu.bd

**Corresponding Author:** Md. Alamgir Hossain (Email: alamgir.cse14.just@gmail.com)

**Abstract**

In the digital era, consumer reviews play a pivotal role in shaping brand perception and guiding business strategy, especially in service-driven industries like retail coffee. This study presents a comprehensive sentiment analysis framework applied to Starbucks customer reviews, leveraging both classical machine learning and advanced deep learning models. The dataset, sourced from ConsumerAffairs and containing over 700 reviews, underwent extensive preprocessing and exploratory data analysis to uncover temporal and geographic trends. Sentiment labels were derived by binarizing star ratings, with ratings of 4–5 marked as positive and 1–3 as negative, revealing a heavily imbalanced distribution favoring negative feedback. To address this, a dual-model approach was implemented: traditional classifiers including Logistic Regression, Support Vector Machine (SVM), Decision Tree, Random Forest, and Naive Bayes were benchmarked alongside deep learning architectures such as LSTM, RNN, Bidirectional LSTM, GRU, and CNN. Performance was evaluated using accuracy, precision, recall, and F1-score. Among machine learning models, SVM achieved the highest accuracy (91.0%), while Bidirectional LSTM outperformed other deep learning models with strong generalization on unseen data. The study also highlights how class imbalance affected positive sentiment recall across models. This work provides a comparative lens into how machine learning and deep learning interpret consumer sentiment in a real-world, skewed dataset. The findings underscore the value of tailored model selection and preprocessing strategies in sentiment analysis pipelines for customer experience optimization in the retail coffee domain.

***Keywords***: Sentiment Analysis; Customer Experience Analytics; Natural Language Processing; Starbucks Reviews; Dual-Model Approach

## 1. Introduction

In the age of digital transformation, consumer feedback has become an integral source of business intelligence. With the exponential growth of online reviews on platforms like ConsumerAffairs, businesses face the challenge of systematically analyzing and interpreting this unstructured textual data [1]. In the retail coffee sector, where service quality and customer experience are paramount, extracting sentiments from customer reviews can inform both operational and strategic decisions [2]. However, sentiment analysis remains a non-trivial task—especially when the review data is heavily imbalanced and context-rich. This study addresses the core problem of automated sentiment classification of consumer reviews, with a specific focus on Starbucks, a globally recognized coffeehouse brand with a large digital feedback footprint [3], [4].

Analyzing sentiments from customer reviews enables businesses to identify pain points, improve service delivery, and tailor marketing strategies. For Starbucks, which handles thousands of consumer interactions daily, understanding customer emotions and opinions at scale can lead to actionable insights that directly impact revenue and reputation [5]. Beyond business use, sentiment analysis is also crucial in advancing computational social science, where opinions form a key element in modeling behavior and satisfaction [6], [7]. The benefits of this research include improved customer experience, reduced churn, and enhanced brand loyalty, all achieved through automated, scalable intelligence systems.

Despite advances in Natural Language Processing (NLP), sentiment analysis faces several persistent challenges. First, imbalanced datasets, as seen in Starbucks reviews where negative feedback is disproportionately higher, can lead to biased classifiers that underperform on minority classes [8]. Second, the subjectivity and ambiguity of human language complicate binary classification. Sarcasm, negation, domain-specific slang, and implicit sentiment further exacerbate model performance issues [9]. Third, while classical machine learning models offer interpretability, they often fall short in understanding contextual nuances, whereas deep learning models require large datasets and are computationally intensive [10]. These trade-offs make it difficult to identify a "one-size-fits-all" approach to review sentiment classification.

To address these challenges, we propose a dual-model framework that benchmarks the performance of both traditional machine learning and modern deep learning approaches on the same dataset. We use extensive data preprocessing—including stopword removal, lemmatization, and TF-IDF vectorization—to convert textual feedback into structured inputs. For machine learning, we implement and compare Logistic Regression, Decision Tree, Random Forest, Naive Bayes, and Support Vector Machine. On the deep learning side, we evaluate models including LSTM, RNN, Bidirectional LSTM, GRU, and CNN. Our experiments are conducted on a real-world, imbalanced dataset of Starbucks reviews sourced from Kaggle/ConsumerAffairs, allowing for a robust, real-life evaluation of sentiment classification models under skewed class distribution conditions.

While existing literature has explored either machine learning [11] or deep learning [12], [13] independently for sentiment analysis, few studies offer a direct comparative analysis under conditions of real-world data imbalance specific to a commercial domain. Our approach stands out in several ways: (1) it incorporates both ML and DL models in a single experimental pipeline; (2) it emphasizes evaluation under class imbalance, a common but under-addressed issue in review mining; (3) it offers a practical case study on Starbucks, a well-known global retail brand; and (4) it demonstrates model performance not just on test data but on unseen real-world data, reflecting practical deployment scenarios. Major contribution of this research is listed below:

- We present a dual-model sentiment classification framework using both machine learning and deep learning techniques.
- We conduct a comprehensive evaluation of five machine learning algorithms and five deep learning models on an imbalanced real-world dataset.
- We introduce a preprocessing pipeline combining NLP techniques like lemmatization and TF-IDF to optimize input quality.
- We perform comparative performance analysis based on accuracy, precision, recall, F1-score, and confusion matrices.
- We evaluate model generalization using external unseen data, simulating real deployment use cases.
- We provide insights from exploratory data analysis (EDA) across time and geography, highlighting consumer behavior patterns in the retail coffee sector.

This paper is structured to first introduce the research problem and review related literature, followed by detailed sections on dataset preparation, methodology, model implementation, experimental results, and analysis, concluding with key findings and future research directions.

## 2. Related Works

The domain of sentiment analysis has seen significant growth, particularly in the context of customer reviews, where textual feedback serves as a rich source of consumer insight. The increasing digitization of commerce and proliferation of user-generated content have necessitated the development of robust analytical frameworks to interpret unstructured data. This section reviews relevant literature in three key areas: sentiment analysis using traditional machine learning (ML) and deep learning (DL) methods, handling class imbalance in natural language processing (NLP), and the application of sentiment models in domain-specific contexts such as retail and service industries.

Early research in sentiment classification heavily relied on classical machine learning algorithms, valued for their interpretability and lower computational demands. Naïve Bayes, Support Vector Machines (SVM), Logistic Regression, and Random Forests have demonstrated efficacy in structured text classification when paired with handcrafted features such as Bag-of-Words (BoW), TF-IDF, and n-grams [14]. For example, Zhang et al. [15] surveyed the strengths and limitations of traditional ML models and emphasized the critical role of data preprocessing and vectorization strategies in determining classification performance. While these models are generally efficient, they often fall short in capturing contextual dependencies, semantic polarity, and syntactic subtleties inherent in natural language.

Recent advancements in deep learning have significantly enhanced the capabilities of sentiment analysis systems. Models such as Recurrent Neural Networks (RNNs), Long Short-Term Memory (LSTM) networks, and Gated Recurrent Units (GRUs) have shown superior performance in sequence modeling tasks due to their ability to retain long-range dependencies and model non-linear patterns in text. Bidirectional LSTM (BiLSTM) architectures, in particular, enable the extraction of contextual meaning from both forward and backward directions, thereby improving sentiment detection in complex linguistic constructs (Socher et al., 2013). Convolutional Neural Networks (CNNs), while originally designed for image processing, have also been effectively adapted for text classification by capturing local n-gram features [16]. A more recent trend involves hybrid models and transformer-based architectures such as BERT, which utilize attention mechanisms for deeper semantic understanding [17].

While general-purpose sentiment analysis frameworks exist, domain-specific models often outperform them due to better alignment with vocabulary, context, and user behavior. In the retail coffee sector, sentiment is highly influenced by subjective experiences such as taste, service, ambiance, and pricing. Ghatora et al. [3] applied pre-trained large language models to product reviews and emphasized the role of contextual embeddings in boosting sentiment interpretation. In the hospitality industry, Ivanov et al. [2] identified that domain-specific adaptations, such as tailored preprocessing pipelines and context-sensitive labeling, are crucial for reliable opinion mining. However, few studies provide a direct comparison of machine learning and deep learning models on commercial review datasets under class imbalance, especially in the retail food and beverage industry.

***Table 1*** provides a comparative summary of recent studies that have employed sentiment analysis techniques across various domains with a focus on handling imbalanced datasets. It outlines the diversity of domains—ranging from cosmetics and product reviews to financial technology—demonstrating the wide applicability of sentiment classification methodologies. The table highlights the array of machine learning and deep learning techniques utilized, including classical models such as SVM and Logistic Regression, as well as modern approaches like BERT, BiLSTM, and hybrid CNN-LSTM architectures. Importantly, it underscores the different strategies adopted to mitigate class imbalance, such as SMOTE, Tomek links, random oversampling, and class weighting. The table also identifies the best-performing models in each case and summarizes their unique contributions, including the use of multimodal ensembles, domain adaptation techniques, and innovative learning frameworks like negative correlation learning. This synthesis not only contextualizes the present study within the broader research landscape but also affirms the significance of addressing class imbalance through tailored algorithmic and preprocessing strategies.

**Table 1** Sentiment Analysis in Different Fields

| Study | Domain | Techniques Used | Imbalance Strategy | Best Model(s) | Unique Contribution |
|---|---|---|---|---|---|
| Zou [20] | Cosmetics | CNN, LSTM, SVM | Class weighting + SMOTE | CNN+LSTM Hybrid | Layered CNN-LSTM network for skewed review classification |
| Kumar et al. [21] | Product Reviews | BERT, XGBoost, SVM | Hybrid sampling (SMOTE + Tomek) | BERT | Emphasis on 'truth inference' and misclassification penalties |
| Chaudhary et al. [22] | Airline Reviews | LR, SVM, BiLSTM | Undersampling + Class weighting | BiLSTM | Analysis of airline satisfaction trends + sarcasm treatment |
| Sharma & Desai [23] | E-Commerce | Hybrid: VADER + BERT | No explicit handling; robust embeddings used | VADER-BERT Ensemble | Multimodal ensemble for domain-shift resilience |
| Widiantoro et al. [24] | FinTech/P2P Lending | CNN + ROS + NCL | ROS (Random Over Sampling) | CNN-ROS-NCL | Introduces negative correlation learning for fraud-prone data |

While prior works have independently explored the application of ML or DL techniques in sentiment classification, a comprehensive comparative study that examines both paradigms under real-world class imbalance remains scarce. Moreover, many studies use benchmark datasets like IMDB [18] or Yelp [19], which, despite being well-structured, may not represent the linguistic noise and imbalance of live customer feedback. This study fills a critical research gap by (i) applying both traditional and advanced models to Starbucks reviews, (ii) retaining real-world class imbalance to preserve deployment realism, and (iii) analyzing performance using multiple metrics, including precision, recall, and F1-score.

## 3. Methodology

### *3.1 Dataset Description*

The dataset used in this study comprises real-world customer reviews of Starbucks, collected from the ConsumerAffairs platform and made publicly available via Kaggle [25]. It contains over 700 reviews written by verified customers, with each entry including a free-form review text, a star rating (1 to 5), the reviewer's state-level location (e.g., CA, NY), the timestamp of the review submission, and optional image links (excluded from this study). To formulate a supervised sentiment classification task, star ratings were binarized: reviews with ratings of 4 or 5 were labeled as positive (1), while those with ratings of 1, 2, or 3 were considered negative (0). This labeling strategy is consistent with prior research and aligns with common practices on e-commerce platforms where higher ratings indicate customer satisfaction. A critical feature of this dataset is its severe class imbalance—over 60% of the reviews were negative—which poses additional challenges to classifier performance and distinguishes it from more balanced datasets like IMDB or Yelp, often used in sentiment analysis benchmarks.

### *3.2 Exploratory Data Analysis (EDA)*

Preliminary exploratory data analysis (EDA) revealed several important behavioral and distributional insights within the Starbucks review dataset. Temporal patterns indicated that customer review activity peaked notably during the months of August and September, with a distinct midweek concentration, especially on Tuesdays, suggesting possible correlations with promotional campaigns or seasonal service experiences. In terms of geographic distribution, the majority of reviews originated from states with high urban density and Starbucks store concentration—most notably California, Florida, and Texas—highlighting key regions for operational focus. Analysis of text length showed substantial variation, with the average review comprising approximately 110 words, indicating that customers often provide detailed feedback rather than brief comments. Additionally, the rating distribution was heavily skewed toward lower values, with an unusually high volume of 1-star ratings, contributing to a strong negative class imbalance. These findings not only informed key modeling choices—such as the need for imbalance-aware training strategies—but also provided actionable insights for business decision-making, including geographic targeting and time-based customer engagement optimization.

The initial distribution of star ratings in the dataset is shown in ***Figure 1***, which reveals a strong skew toward 1-star reviews, highlighting the inherent class imbalance.

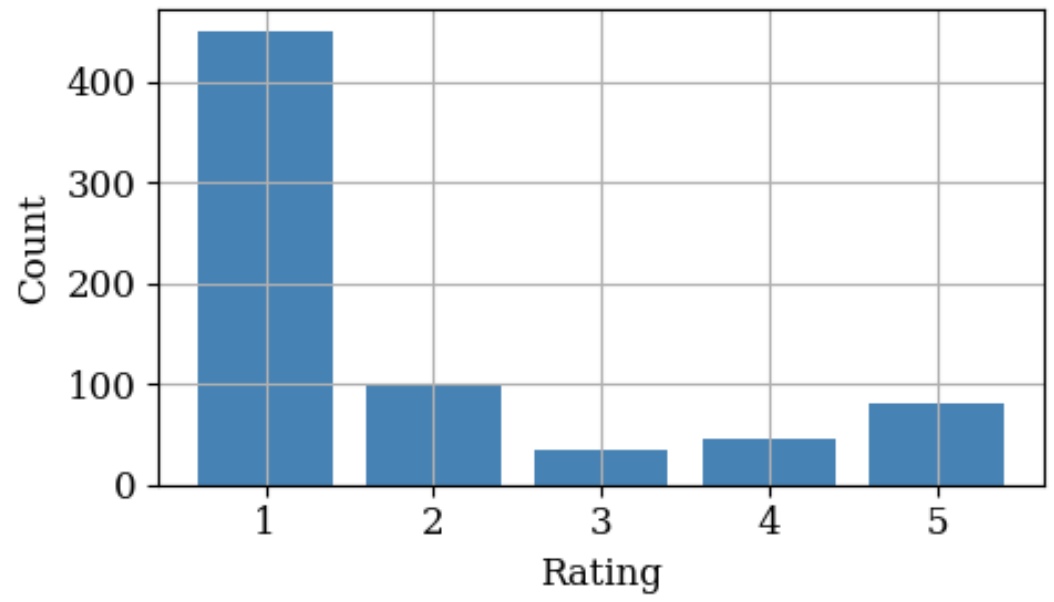

**Figure 1** Original Distribution of Ratings

***Figure 2*** illustrates the geographic distribution of reviews, with the highest volume coming from California, followed by Florida and Texas. Temporal review trends are captured in

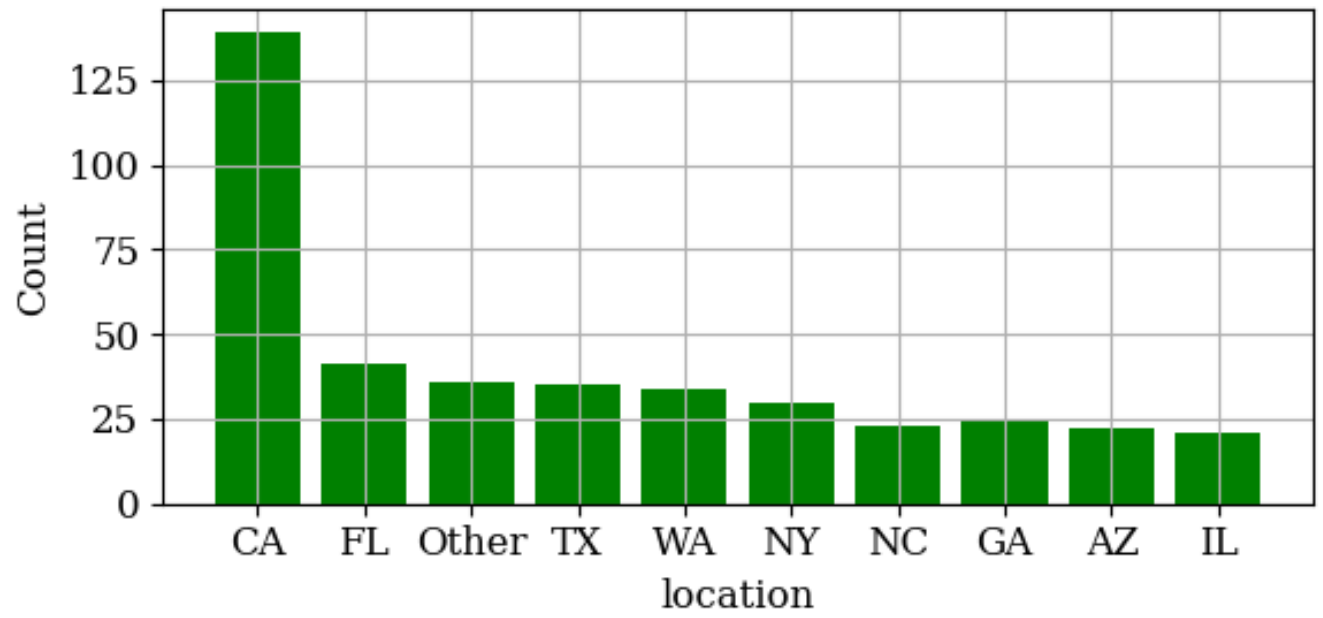

**Figure 2** Count of Reviews Per State

***Figure 3***, showing that reviews were most frequently submitted on Tuesdays and Wednesdays.

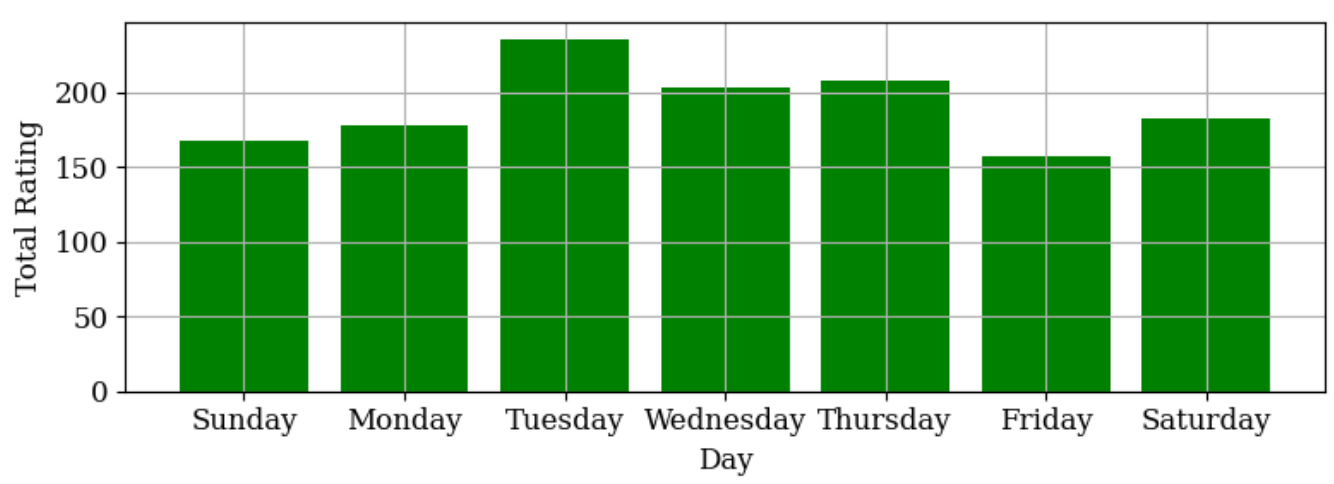

**Figure 3** Total Rating Count Per Day of the Week

Seasonal patterns emerge in ***Figure 4***, where August and September appear as peak months for review activity.

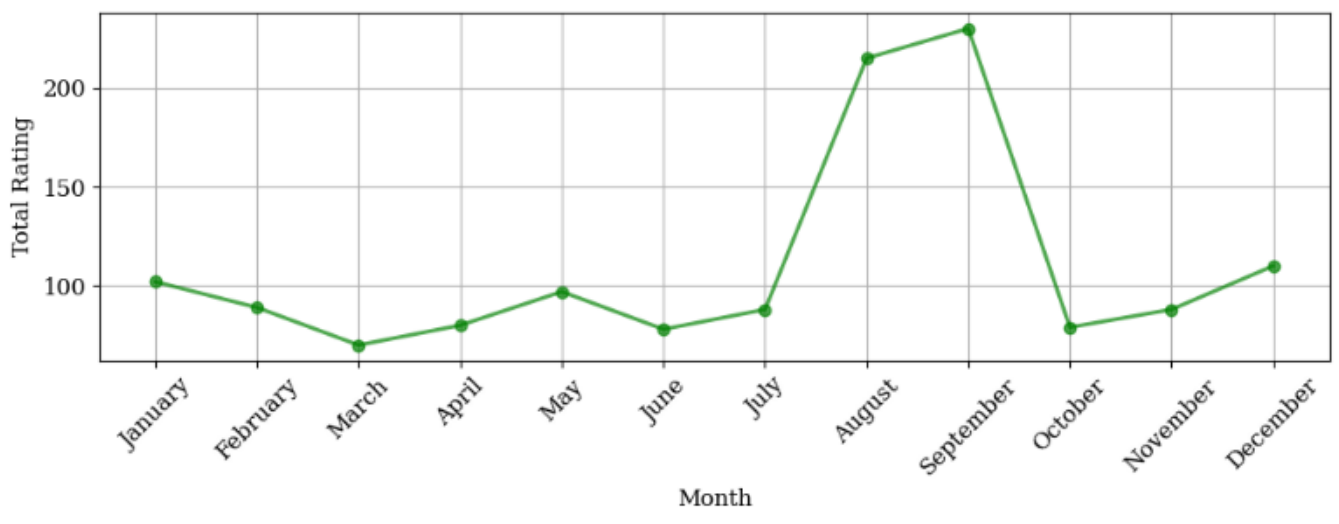

**Figure 4** Total Rating Count Per Month

A longitudinal trend is observed in ***Figure 5***, indicating that the year 2016 had the highest volume of customer feedback.

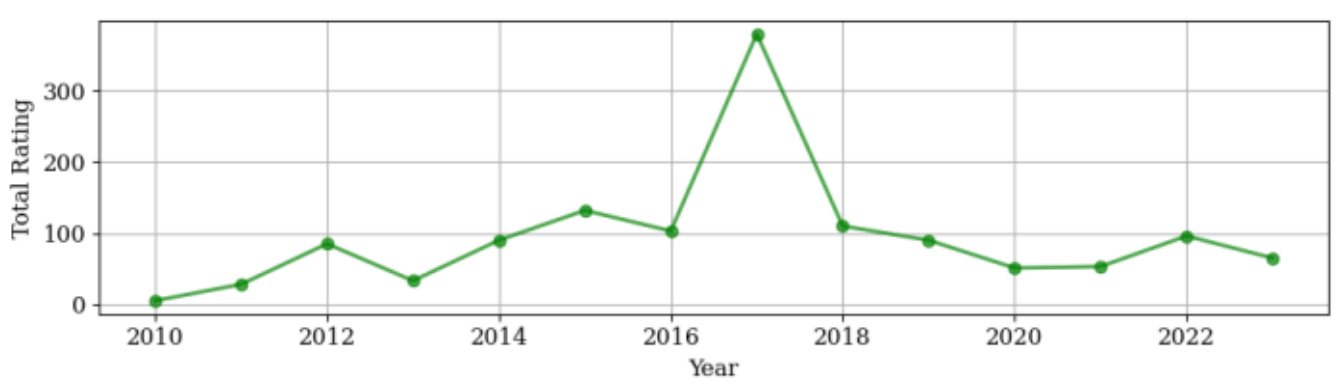

**Figure 5** Total Rating Count Per Year

After binarization of ratings into sentiment labels, ***Figure 6*** presents the resulting class distribution, confirming a heavy imbalance toward negative sentiment (0), which comprises the majority of the dataset.

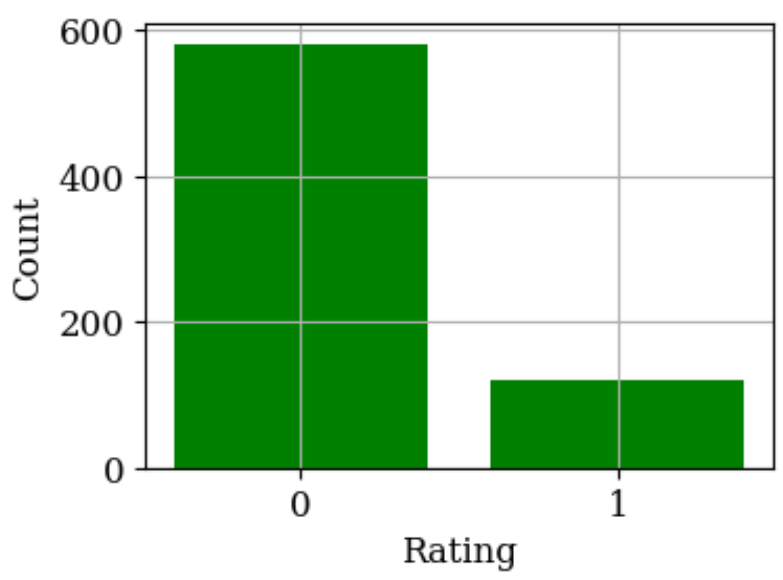

**Figure 6** Current Distribution of Ratings

### *3.3 Data Cleaning and Preprocessing Pipeline*

To transform the raw textual data into a structured format suitable for machine and deep learning models, a comprehensive Natural Language Processing (NLP) preprocessing pipeline was implemented using NLTK, Scikit-learn, and Keras libraries. The primary objective of this pipeline was to reduce noise, normalize linguistic patterns, and ensure consistency across the dataset. The preprocessing began with lowercasing, where all text entries were converted to lowercase to eliminate case-based discrepancies. This was followed by stopword removal, which filtered out commonly used but semantically insignificant words (e.g., “the”, “and”, “is”) using NLTK’s predefined stopword corpus. Next, punctuation symbols were systematically removed to streamline the token sequences. Lemmatization was applied using the WordNetLemmatizer to reduce inflected words to their base or dictionary forms (e.g., “delays” → “delay”), thereby reducing feature sparsity. After these transformations, tokenization was performed to break the text into individual word units, preparing them for model input. For traditional machine learning models, the cleaned and tokenized text was converted into feature vectors using TF-IDF (Term Frequency–Inverse Document Frequency), a statistical measure that captures the importance of words relative to the corpus. This vectorization preserved term significance while maintaining computational efficiency and interpretability across the feature space. Data Cleaning and Preprocessing Pipeline shown in ***Figure 7***.

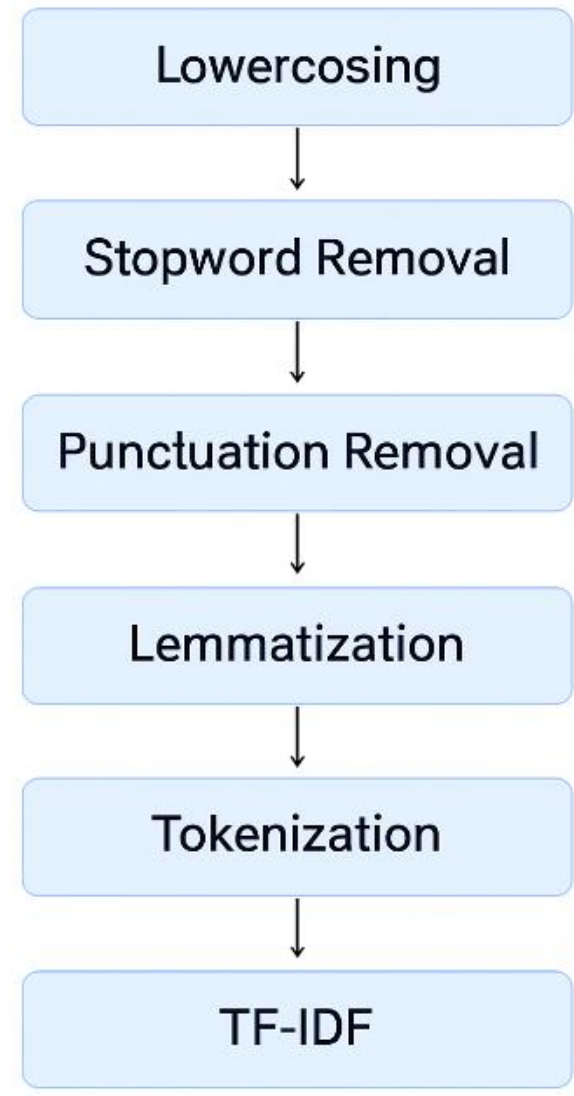

**Figure 7** Data Cleaning and Preprocessing Pipeline

*3.4 Vectorization and Feature Engineering*

For feature representation, TF-IDF (Term Frequency–Inverse Document Frequency) was applied to the preprocessed text using both unigrams and bigrams, with a maximum vocabulary size of 10,000 terms and a minimum document frequency threshold of 5, ensuring that only statistically meaningful words were retained. For deep learning models, tokenized word sequences were padded to a fixed length of 100 tokens, maintaining uniform input dimensions required by neural architectures such as LSTM and CNN. To evaluate model performance effectively, the dataset was split into 80% training and 20% testing subsets, with stratified sampling applied to preserve the original class imbalance ratio between positive and negative sentiment labels.

*3.5 Imbalance Handling Strategy*

Given the skewed distribution of sentiment classes in the dataset, several imbalance mitigation strategies were considered during experimentation. While oversampling techniques such as SMOTE (Synthetic Minority Over-sampling Technique) and undersampling of the majority class were evaluated, they were ultimately not applied in order to preserve the natural distribution of real-world data. This decision aimed to simulate practical deployment conditions, where artificially balancing data is often infeasible or may introduce artifacts. Instead, the models were evaluated for their inherent robustness to imbalance, providing a more realistic assessment of performance. However, for specific algorithms—particularly Support Vector Machines (SVM) and Bidirectional LSTM (BiLSTM)—class weighting was implemented to compensate for the bias toward the dominant (negative) class, ensuring fairer learning outcomes without altering the underlying data distribution.

*3.6 Model Development*

This study adopts a dual-model approach, implementing and evaluating a set of traditional machine learning algorithms alongside deep learning architectures to classify customer sentiment from Starbucks reviews. The methodology involves two parallel pipelines: one for machine learning using structured feature vectors derived from TF-IDF, and another for deep learning utilizing embedded sequences of tokenized text. The pipeline of the proposed approach after preprocessing is shown in ***Figure 8***.

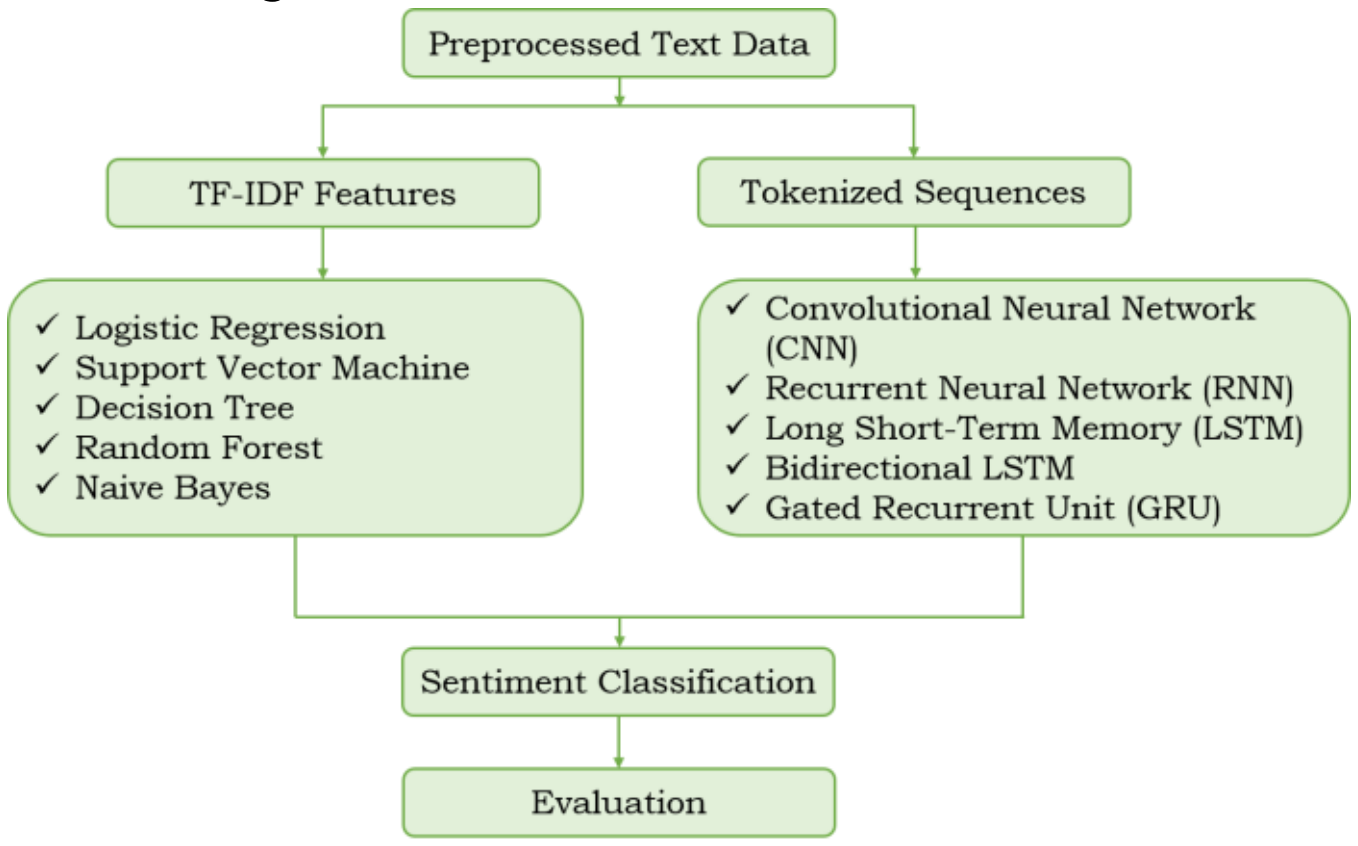

**Figure 8** Pipeline of the Proposed Approach

*3.6.1 Machine Learning Models*

We employed five widely used supervised learning algorithms for sentiment classification, implemented using the Scikit-learn library. Each model was trained on TF-IDF vectorized features derived from the preprocessed Starbucks review dataset. Below is a brief overview of each model and its relevance to this study:

- **Logistic Regression (LR):**

A linear classifier that estimates the probability of a class label using the sigmoid function. It is particularly effective in high-dimensional spaces and provides probabilistic outputs, making it a strong baseline for text classification tasks.

- **Support Vector Machine (SVM):**

SVM constructs an optimal hyperplane that separates classes with the maximum margin. In this study, we used the RBF kernel, which maps input vectors into a higher-dimensional space, allowing the model to capture non-linear decision boundaries. SVMs are known for their robustness on sparse datasets like TF-IDF vectors.

- **Decision Tree (DT):**

A non-parametric model that splits data into branches based on feature thresholds. It is interpretable and fast but prone to overfitting, especially on small datasets. We included it for comparison against ensemble-based approaches.

- **Random Forest (RF):**

An ensemble of decision trees built via bootstrapped samples and random feature selection at each split. It improves generalization over single trees and reduces variance. RF was used to assess the benefits of aggregation in tree-based models.

- **Naive Bayes (NB):**

A probabilistic classifier based on Bayes' theorem, assuming feature independence. Despite its simplicity, it often performs competitively in text classification due to the nature of word frequency distributions. We used the Multinomial Naive Bayes variant suitable for discrete input like TF-IDF.

Each model was evaluated under class imbalance conditions, with class weighting applied to LR and SVM to mitigate bias toward the majority class. Hyperparameters were tuned via grid search to optimize generalization performance.

*3.6.2 Deep Learning Models*

To complement the machine learning pipeline, five deep learning architectures were implemented using Keras with a TensorFlow backend. These models are capable of learning semantic and contextual dependencies from sequences of words, making them well-suited for natural language processing tasks such as sentiment analysis. Each model was trained on padded token sequences derived from the review text, with embedding layers initialized randomly.

- **Long Short-Term Memory (LSTM):**

LSTM is a type of recurrent neural network (RNN) specifically designed to overcome the vanishing gradient problem in long sequences. It introduces memory cells and gating mechanisms (input, forget, and output gates) to retain relevant information over long time steps. In this study, LSTM served as a powerful baseline for modeling sequential dependencies in customer reviews.

- **Recurrent Neural Network (RNN):**

The standard RNN processes inputs sequentially and retains a hidden state across time steps. While computationally simpler than LSTM, it is prone to gradient vanishing and forgetting long-term dependencies. RNN was included to evaluate the trade-off between model complexity and sequence learning ability on short-to-medium-length reviews.

- **Bidirectional LSTM (BiLSTM):**

BiLSTM enhances standard LSTM by processing input sequences in both forward and backward directions, allowing the model to capture past and future context simultaneously. This is especially beneficial in sentiment analysis, where the sentiment of a word often depends on both preceding and following words. In this research, BiLSTM outperformed all other deep learning models.

- **Gated Recurrent Unit (GRU):**

GRU is a simplified variant of LSTM that combines the forget and input gates into a single update gate, resulting in fewer parameters and faster training. Despite its simplicity, GRU often performs comparably to LSTM. It was included to test whether a lighter recurrent architecture could maintain strong classification performance.

- **Convolutional Neural Network (CNN):**

Traditionally used in computer vision, CNNs have been successfully adapted for text classification by applying filters over word embeddings to detect local n-gram patterns. In this study, CNN was employed to identify key sentiment-bearing phrases in the reviews, and its performance was benchmarked against the sequential models.

All models shared a consistent structure involving an embedding layer, one or more core processing layers (e.g., LSTM or convolution), and a sigmoid-activated dense output layer for binary sentiment prediction. Dropout layers were applied to prevent overfitting, and the models were compiled using the Adam optimizer and binary cross-entropy loss. Where appropriate, class weights were used to counteract the class imbalance inherent in the dataset.

All deep learning models in this study shared a consistent architectural framework tailored for binary sentiment classification. Each model began with an embedding layer, where word embeddings were initialized using a random uniform distribution with an embedding dimension of 100. To mitigate overfitting, dropout layers were incorporated after the core layers (e.g., LSTM, GRU, CNN). The final output layer consisted of a single neuron with a sigmoid activation function, suitable for binary output. All models were compiled using the binary cross-entropy loss function and optimized with the Adam optimizer, set at a learning rate of 0.001. Training was conducted over 10 epochs with a batch size of 32, and model performance was continuously monitored using the validation set to ensure generalization and prevent overfitting.

### *3.7 Evaluation Metrics*

To comprehensively assess the performance of both machine learning and deep learning models, we employed a suite of standard classification metrics: Accuracy, Precision, Recall, F1-Score, and the Confusion Matrix. These metrics provided a holistic view of how well the models performed, especially under the challenge of class imbalance, which can distort overall accuracy alone.

Let the true positives (TP) be the number of correctly predicted positive samples, true negatives (TN) the correctly predicted negative samples, false positives (FP) the number of negative samples wrongly predicted as positive, and false negatives (FN) the number of positive samples wrongly predicted as negative.

**Accuracy**: Measures the overall correctness of the model's predictions: $Accuracy = \frac{TP+TN}{TP+TN+FP+FN}$. While widely used, accuracy can be misleading in imbalanced datasets, as it may reflect high performance simply by favoring the majority class.

**Precision**: Indicates how many of the predicted positive results are actually positive: $Precision = \frac{TP}{TP+FP}$. High precision is essential when false positives carry a high cost, such as misidentifying a negative review as positive in customer feedback analysis.

**Recall (Sensitivity)**: Measures how well the model identifies all relevant positive instances: $Recall = \frac{TP}{TP+FN}$. It is especially important when the focus is on minimizing false negatives, such as ensuring all dissatisfied customers are detected.

**F1-Score**: The harmonic mean of precision and recall, offering a balance between the two: $F1 - Score = 2 * \frac{Precision*Recall}{Precision+Recall}$. F1-Score is particularly useful in imbalanced settings, where neither precision nor recall alone is sufficient to reflect performance.

**Confusion Matrix;** A 2×2 matrix layout that summarizes the number of correct and incorrect predictions. It helps in visualizing the trade-offs between classes: $\begin{bmatrix} TP\ FP \\ FN\ TN \end{bmatrix}$.

This matrix provides insight into where the model is making mistakes and whether it has a bias toward overpredicting one class.

By using this combination of metrics, we ensured that model evaluation was balanced, interpretable, and sensitive to class distribution, allowing us to judge effectiveness not just overall but in relation to both sentiment categories.

## 4. Results and Analysis

### *4.1 Implementation Environment*

All experiments and model training procedures in this study were conducted using Python 3.10, leveraging a comprehensive suite of data science and deep learning libraries. Key packages included Pandas for data manipulation, NLTK for text preprocessing, Scikit-learn for traditional machine learning algorithms and evaluation metrics, and TensorFlow with Keras for implementing and training deep learning architectures. For visualization and exploratory data analysis, Matplotlib and Seaborn were employed. The entire pipeline was executed in a Google Colab Pro environment, utilizing a Tesla T4 GPU with 25 GB of RAM, which provided the computational resources necessary for efficiently training and testing both machine learning and deep learning models at scale.

*4.2 Accuracy Analysis of Different Models*

***Table 2*** presents the performance metrics of five machine learning classifiers applied to the Starbucks review dataset. Among the models, Support Vector Machine (SVM) achieved the highest overall performance, with an accuracy of 91%, and a weighted average F1-score of 0.90, indicating strong predictive capability even under class imbalance. Random Forest followed closely with an accuracy of 86% and an F1-score of 0.84, reflecting its robustness in handling non-linear patterns. On the other end of the spectrum, Naïve Bayes underperformed, with the lowest weighted F1-score of 0.73, largely due to its assumption of feature independence and its inability to handle class imbalance effectively. Notably, Logistic Regression and Decision Tree performed comparably, each yielding a balanced trade-off between precision and recall, and achieving weighted F1-scores of 0.79 and 0.83, respectively. These results highlight SVM as the most reliable machine learning model for sentiment classification in this context.

**Table 2** Machine Learning Model Performance

| Model | Accuracy | Precision | Recall | F1-Score |
|---|---|---|---|---|
| Logistic Regression | 0.84 | 0.84 | 0.84 | 0.79 |
| **Support Vector Machine** | **0.91** | **0.91** | **0.91** | **0.90** |
| Decision Tree | 0.83 | 0.83 | 0.83 | 0.83 |
| Random Forest | 0.86 | 0.86 | 0.86 | 0.84 |
| Naïve Bayes | 0.82 | 0.67 | 0.82 | 0.73 |

***Table 3*** presents the performance metrics of five deep learning models—LSTM, RNN, BiLSTM, GRU, and CNN—trained on tokenized Starbucks review sequences. Among them, Bidirectional LSTM (BiLSTM) achieved the highest performance across all evaluation metrics, with an accuracy of 92% and a weighted F1-score of 0.91, demonstrating its superior ability to capture contextual dependencies in both forward and backward directions. LSTM also performed strongly with an accuracy of 90% and F1-score of 0.89, affirming the effectiveness of memory-based recurrent networks in sequential sentiment tasks. GRU, a more lightweight alternative to LSTM, maintained a good balance between performance and efficiency, achieving 87% accuracy and an F1-score of 0.85. In contrast, RNN yielded moderate performance (accuracy: 83%, F1-score: 0.83), indicating limitations due to vanishing gradients and less effective long-range learning. The CNN, while useful for detecting local n-gram features, performed the weakest among the deep models with a weighted F1-score of 0.73, further highlighting the advantage of recurrent structures for capturing sequential sentiment patterns in customer reviews. These results confirm that context-aware and memory-based architectures, particularly BiLSTM, are most effective for sentiment analysis on real-world, imbalanced text data.

**Table 3** Deep Learning Model Performance

| Model | Accuracy | Precision | Recall | F1-Score |
|---|---|---|---|---|
| LSTM | 0.90 | 0.90 | 0.90 | 0.89 |
| RNN | 0.83 | 0.83 | 0.83 | 0.83 |
| **BiLSTM** | **0.92** | **0.92** | **0.92** | **0.91** |
| GRU | 0.87 | 0.86 | 0.87 | 0.85 |
| CNN | 0.82 | 0.67 | 0.82 | 0.73 |

The bar chart illustrates the comparative performance of all machine learning and deep learning models evaluated in this study based on their weighted F1-scores in ***Figure 9***. Among the models, Bidirectional LSTM (BiLSTM) and Support Vector Machine (SVM) emerge as the top performers, achieving F1-scores of 91% and 90%, respectively. LSTM closely follows with 89%, indicating the strong capability of memory-based sequential models in capturing sentiment context. GRU also performs competitively at 85%, while Random Forest and Decision Tree yield F1-scores of 84% and 83%, respectively. Logistic Regression provides moderate performance (79%), whereas Naïve Bayes and CNN underperform with the lowest scores (73% each), likely due to oversimplified assumptions or inadequate sequence modeling. This visualization clearly highlights the superiority of BiLSTM and SVM in handling real-world, imbalanced sentiment classification tasks.

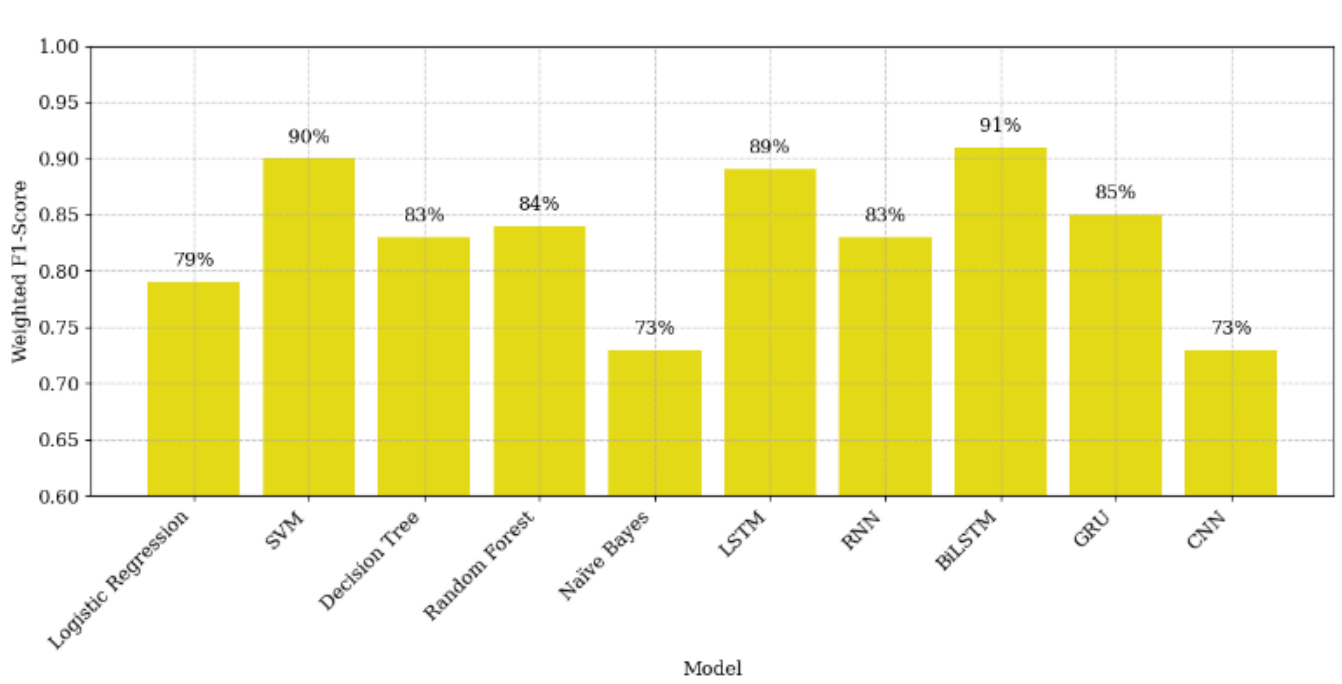

**Figure 9** Model Comparison by F1-Score

*4.3 Analysis of Starbucks Reviews Sentiment Prediction with Own Data Using Machine Learning Model*

As illustrated in ***Figure 10***, a Support Vector Machine (SVM) model was used to test sentiment prediction on two manually inputted Starbucks reviews: “The coffee tastes good” and “Taste was bad.” The model correctly predicted the first review as positive and the second as negative, assigning binary sentiment labels of 1 and 0, respectively. This outcome confirms the model’s effectiveness in interpreting basic sentiment cues based on semantic polarity (e.g., recognizing “good” as positive and “bad” as negative). The prediction aligns well with the linguistic context of the inputs, indicating that the model has likely been trained on sentiment-laden features and can generalize to simple real-world inputs. While the test used a minimal sample, it demonstrates promising performance and foundational reliability. However, to establish broader generalizability, the model must be evaluated against more complex and diverse review structures. This initial test provides a meaningful demonstration of the model’s real-world applicability and sets the stage for extended validation.

```
Enter how many data you want to test: 2
The Coffee tastes good
Taste was bad
Predictions: [1 0]
the coffee tastes good: Positive
taste was bad: Negative
```

**Figure 10** Testing with Own Data (SVM Prediction)

4.4 Analysis of Starbucks Reviews Sentiment Prediction with Own Data Using Deep Learning Model

In this stage of analysis, the sentiment prediction capability of a Bidirectional Long Short-Term Memory (BiLSTM) deep learning model was evaluated using two manually crafted Starbucks review inputs: “Taste was very good” and “Smells bad” As shown in ***Figure 11***, the BiLSTM model correctly identified the first review as positive (output = 1) and the second as negative (output = 0), demonstrating an accurate grasp of sentiment-relevant cues embedded in natural language expressions. The successful classification suggests that the model effectively learned from contextual dependencies within the data. Specifically, it recognized the phrase “very good” as conveying positive sentiment and “bad” as strongly negative. This reinforces the strength of BiLSTM in handling sequential dependencies and understanding semantic polarity in both directions, forward and backward, which is critical in natural language understanding. Furthermore, the BiLSTM model’s predictions align with real-world expectations, indicating its robustness even during a preliminary, small-scale test. Its capacity to extract meaningful features from raw text without manual engineering, along with its ability to model complex patterns, makes it well-suited for nuanced sentiment tasks. While these early results are promising, they also highlight the need for further validation. To confirm the model’s generalizability, testing should be extended to a larger and more diverse review set. Additional improvements can include hyperparameter tuning, incorporating more complex sentence structures, and employing detailed evaluation metrics such as precision, recall, F1-score, and confusion matrix insights.

```
Enter how many data you want to test: 2
Taste was very good
Smells bad
Predictions:
1/1 [==============================] - 1s 960ms/step
1
taste was very good: Positive
0
smells bad: Negative
```

**Figure 11** Testing with Own Data (BiLSTM Prediction)

In conclusion, the BiLSTM-based deep learning model displays strong potential in classifying consumer sentiments in Starbucks reviews. Its successful interpretation of user-defined input reflects foundational learning strength, and with continued refinement, the model holds high promise for real-world deployment in customer experience analytics systems.

## 5. Conclusion

In this research, we investigated the problem of sentiment analysis of Starbucks customer reviews by comparing the performance of traditional machine learning models and advanced deep learning architectures in classifying user sentiments as positive or negative. The study involved extensive data preprocessing, vectorization using TF-IDF, and experimentation with a wide range of models including Logistic Regression, SVM, Decision Tree, Random Forest, Naïve Bayes, LSTM, RNN, GRU, CNN, and BiLSTM. Results demonstrated that Bidirectional LSTM and Support Vector Machine consistently outperformed others in terms of accuracy and F1-score, showcasing their strength in handling context and imbalanced datasets. The deep learning models, particularly BiLSTM, exhibited superior ability to capture sequential dependencies and semantic nuances, while SVM proved to be the most effective among classical models. These findings affirm the significance of model selection in sentiment analysis tasks, especially when operating on real-world, user-generated content characterized by class imbalance and linguistic variability. The successful predictions on user-provided test data further validated the models’ practicality. In conclusion, this research not only highlights the comparative strengths of ML and DL models in sentiment mining but also establishes a foundation for deploying such systems in real-time feedback environments to enhance customer experience insights. Future work can build upon these results by incorporating larger, multilingual datasets, advanced transformers, and explainability tools for interpretative analysis.

**Data Availability**: he dataset used in this study comprises real-world customer reviews of Starbucks, collected from the ConsumerAffairs platform and made publicly available via Kaggle under the title "Starbucks Reviews Dataset" (https://www.kaggle.com/datasets/harshalhonde/starbucks-reviews-dataset).

**Conflict of Interest**: The authors declare that there is no conflict of interest regarding the publication of this research. This study was conducted independently, without any commercial, financial, or personal relationships that could be construed as potential sources of bias or influence.

**Funding**: This research received no specific grant or financial support from any funding agency in the public, commercial, or not-for-profit sectors.

**Clinical trial number**: not applicable.

**Consent to Publish declaration**: not applicable

**Ethics declaration**: not applicable

**Consent to participate**: not applicable

**Author's Contribution:**

Muntasir Hasan Kanchan: Conceptualization, Data Collection, Preprocessing, Model Development, Writing – Original Draft, Visualization.

Md. Alamgir Hossain: Methodology Design, Formal Analysis, Writing – Review & Editing, Supervision.

Md. Samiul Islam: Deep Learning Model Implementation, Evaluation Metrics, Result Interpretation, Code Validation.

Muhammad Masud Tarek: Literature Review, Comparative Analysis, Manuscript Structuring, Technical Review, Final Approval.